# Incremental Outlier Detection Modelling for Fraud Detection in Finance and Health Care

Vivek Yelleti[1*], Chaduvula Priyanka[2]

[1]Department of Computer Science & Engineering, National Institute of Technology Warangal, Hanumkonda, Warangal – 506004, Telangana
[2]Jawaharlal Nehru Technological University – Gurajada Vizianagaram (JNTU-GV), Vizianagaram, 535003, Andhra Pradesh

vivek.yelleti@gmail.com[1] ; priyankachaduvula510@gmail.com[2]

**Abstract**

In the era of real-time data, traditional methods often struggle to keep pace with the dynamic nature of streaming environments. In this paper, we proposed a hybrid framework where in (i) stage-I follows a traditional approach where the model is built once and evaluated in a real-time environment, and (ii) stage-II employs an incremental learning approach where the model is continuously retrained as new data arrives, enabling it to adapt and stay up to date. To implement these frameworks, we employed 8 distinct state-of-the-art outlier detection models, including one-class support vector machine (OCSVM), isolation forest adaptive sliding window approach (IForest ASD), exact storm (ES), angle-based outlier detection (ABOD), local outlier factor (LOF), Kitsune's online algorithm (KitNet), and K-nearest neighbour conformal density and distance based (KNN CAD). We evaluated the performance of these models across seven financial and healthcare prediction tasks, including credit card fraud detection, churn prediction, Ethereum fraud detection, heart stroke prediction, and diabetes prediction. The results indicate that our proposed incremental learning framework significantly improves performance, particularly on highly imbalanced datasets. Among all models, the IForest ASD model consistently ranked among the top three best-performing models, demonstrating superior effectiveness across various datasets.

## 1. Introduction

Streaming data refers to a continuous influx of information generated through various data sources in real time. Handling such data real-time is crucial for effective decision making and risk mitigation practices, particularly in fields such as medical and financial sectors. In healthcare applications [1], continuous monitoring of patients through ECG signals and smart devices generate a constant flow of vast information. Similarly, in finance sector [2], real-time transaction data, stock market, and other fraud detection systems [3] occupy a pinnacle demand for immediate mitigation measures.

Within these streaming applications, certain data points may exhibit unexpected or anomalous behaviour – these are known as outliers [4]. An outlier [5] is defined as follows "*An observation (or subset of observations) which appears to be inconsistent with the remainder of that set of data*". Detecting such outliers and handling them effectively is very critical. As detecting anomalies behaviour, suppose in healthcare, abnormal heart rhythms could indicate early signs of critical conditions [1], requiring immediate intervention. In finance, suspicious transactions might indicate the occurrence of fraudulent transactions [6], and mitigating such transactions promptly could save millions of dollars and, most importantly, improve the customer-bank relationship.

The major challenges while handling streaming data [7] are as follows: (i) latency, (ii) scalability, and (iii) need to incrementally train the machine learning (ML) model. Traditional ML models often could not capture the evolving stream patterns and require frequent retraining from scratch, which incurs huge computational complexity. Owing to these challenges, incremental learning algorithms came into existence, where the underlying ML model retrains with the new evolving pattern, thereby improving the robustness of the model. By incorporating this principle, we ensure continuous learning and adaptability, and this makes them better suited for real-time healthcare monitoring and fraud detection effectively. Additionally, real-time modelling faces another major challenge, i.e., the cold start problem [8], where the model initially lacks sufficient data to make accurate predictions. Hence, in the initial period, the model makes unreliable predictions and suffers from high misclassification rate. This is particularly noticed post-deployment of the ML models in streaming environments and persists till it encounters adequate patterns to distinguish between normal and anomalous behaviour.

Given these challenges, incremental learning approaches are gaining traction. Unlike traditional models, incremental models can dynamically adapt to new patterns without requiring complete retraining, making them well-suited for real-time applications. This motivation drives the need for a hybrid framework that integrates both offline model building and incremental learning to handle streaming data efficiently in the context of outlier detection. In this study, we propose a hybrid framework consisting of two stages: (i) Stage I, where the model is initially trained using a traditional approach and then deployed for real-time evaluation, and (ii) Stage II, where an incremental learning approach is implemented, allowing the model to continuously update and adapt as and when new data streams arrives.

The major highlights of the project are as follows:

- Developed a methodology for an online incremental outlier detection framework to address financial and healthcare challenges in the streaming environment.
- Analyzed the effectiveness of various outlier detection algorithms for streaming data over several imbalanced datasets.
- Compared the performance of incremental model training with offline model building to demonstrate the superiority of the proposed approach.

The remainder of this paper is organized as follows: Section 2 focuses on the literature review, and Section 3 covers a brief overview of the employed outlier detection techniques. Section 4 discusses the proposed framework for online outlier detection. Section 5 presents the dataset description, environmental setup, discusses the results. Section 6 presents the conclusion and future work.

## 2. Related Works

Degirmenci and Karal [9] proposed incremental outlier detection model (iMCOD), which works as follows, initially the incoming samples are passed to incremental support vector machine (iSVM), and later incremental local outlier factor (iLOF) is employed to compute the outlier score. Here, the distance between the incoming samples with the nearest neighbour which was classified to a particular class. Yang et al. [10] proposed sliding window based outlier detection to prune the unnecessary information by employing information entropy and named it as information entropy pruning multi dimensional outlier detection (IPMOD).

Cai et al. [11] proposed maximal pattern based outlier detection (MFP-OD), which utilizing most frequently occurring patterns to discriminate between the outliers and normal samples. MFP-OD comprises three deviation factors to identify the degree of deviation of each transaction. Gao et al. [12] proposed cube-based outlier detection (CB-ILOF), which integrates ILOF algorithm to dynamically update thus extracted profiles of data points. Initially, the data space is divided into multiple cubes, then the outlier detection of data points is transferred into the outlier detection of cubes.

Xiaolan et al. [13] proposed clustering based approach for outlier detection, where the clusters are incrementally updated as the new network samples arrive. Their approach classified the outliers as global outliers and local outliers based on the global and local densitiy thresholds, respectively. Raut et al. [14] proposed adaptive clustering-based event detection scheme for IoT (AEDS-IoT) stream data, by analysing the patterns of the data distribution. The event patterns are detected by employing adaptive clustering technique. Once the event patterns are identified, then the outliers are labelled based on spatial-temporal conditions.

Ducharlet et al. [15] introduced two novel methods for outlier detection, (i) DyCF, which utilizes christoffel function from the theory of approximation and orthogonal polynomials, and (ii) DyCG, that incorporates the Christoffel function thereby eliminating the requirement for fine tuning parameters. Hu et al. [16] proposed an adaptive stream outlier detection (ASOD) where it incorporated K-nearest neighbour (KNN) is employed to limit the influence of new data locally rather than globally, and Gaussian mixture model (GMM) provides the life cycle and the anomaly index to facilitate the cause analysis.

Pei et al. [17] proposed a novel framework to detect anomalies in traffic environment, where they utilized a subspace streaming model to represent all prior non-anomalous data. They proved that their approach outperformed singular value decomposition (SVD) both in terms of accuracy and time complexity. Alghushairy et al. [18] proposed another anomaly based outlier detection (NODS) where the outlier behaviour is noticed in the incoming traffic stream. They verified the effectiveness of their model on NSL-KDD and CICIDS2017 datasets.

Devarapu et al. [19] proposed outlier detection method to identify the malicious transactions and identify the anomaly behaviour in the transactional patterns. Bah et al. [20] proposed micro cluster with minimal probing (MCMP) which is hybrid approach of micro cluster outlier detection (MCOD) and thresh_LEAP. It adopt micro-clusters to differentiate between strong and trivial inliers.

The key differences between the existing approaches between the proposed is as follows:

- Existing approaches either relied on historical data or trained directly in real-time. In the former case, the training with historical data does not abstract the real-world data shift, whereas, the latter case, introduces the cold start problem. To mitigate these issues, we introduced a hybrid architecture and analyzed the performance over several finance and healthcare problems.

# 3. Overview of the employed Techniques

The following are the techniques that are employed in this study.

### 3.1 One class support vector machine

One class support vector machine (OCSVM) [27] is an unsupervised variant of the support vector machine algorithm (SVM) algorithm, specifically designed for anomaly detection and outlier detection problems. The major distinction between SVM and OCSVM is as follows:

- SVM operates in supervised learning where the label data information is provided and necessary to build the classifier.
- OCSVM primarily learns the decision boundaries of the majority class samples, i.e., normal class samples in a dataset. It predicts the anomalies or outliers when any data points fall outside this learned boundary.

In other words, OCSVM constructs a boundary (or hyperplane) that covers the majority of the data from potential anomalies. The data points that fall within this boundary are considered as normal samples ( or negative class), while the samples that fall outside the hyperplane are classified as outliers (or positive class).

### 3.2 Isolation Forest adaptive sliding window

Isolation forest (IForest) is an ensemble unsupervised outlier detection algorithm that, like random forests, built over several decision trees. However, unlike traditional classification trees, IForest is specifically designed to isolate anomalies efficiently. IForest isolates observations by randomly selecting a feature and then randomly selecting a split value between the maximum and minimum values of the selected feature. The important steps involved in IForest are as follows:

1. A random sub-sample of the data is selected and assigned to a binary tree.
2. Branching of the tree starts by selecting a random feature first. Then branching is done on a random threshold.

3. If the value of a data point is less than the selected threshold, it goes to the left branch or else to the right. Thus a node is split into left and right branches.
4. This process from step 2 is continued recursively till each data point is completely isolated or till max depth (if defined) is reached.
5. The above steps are repeated to construct random binary trees.

To enhance the real-time adaptability of IForest in streaming data environments, IForest is integrated with an adaptive sliding window approach, resulting in IForest ASD [24].

### 3.3 Linear Outlier Factor

Local outlier factor (LOF) is an outlier detection algorithm proposed in ref. [23] for identifying anomalous data points. This operates by measuring their local deviation of a given data point w.r.t its neighbours. The core idea of LOF is based on local density estimation, where the locality of a point is determined by K-nearest neighbours (KNN). The density of a given point is estimated by computing the reachability distance (i.e., Euclidean distance) of its neighbours. By comparing the local densities of a data point and its surrounding neighbors, the algorithms can distinguish regions of similar density from points that exhibit significantly lower density, thereby identifying potential outliers.

### 3.4 Angle based Outlier Detection

Angle based outlier detection (ABOD) proposed by Kriegel et al. [28], was developed to address the limitations of traditional distance based anomaly detection methods, particularly in high dimensional spaces. In high dimensional datasets, comparing distances between data points becomes ineffective due to the curse of dimensionality. To overcome this, ABOD evaluates the variance of angles formed between the difference vectors of a given data point and all other points in the dataset. The working principle of ABOD is explained below:

- Farther point exhibit lower variance in angles.
- Within a cluster, the variance of angles among points is relatively high.
- At the cluster boundaries, the variance of angles decreases.
- For outliers, the variance of angles is significantly small, making them easily distinguishable.

### 3.5 Exact Storm

Exact storm [22] is an algorithm, specifically designed for streaming data and mainly consists of two different procedures, (i) stream manager, and (ii) query manager.

- The stream manager is responsible for receiving incoming data streams, which can arrive either as single data points, or micro-batches.
- Upon receiving new stream data, the algorithm updates its data structure based on the computed summary of the current window.

- The main component of the Exact storm is the indexed stream buffer (ISB), which consists of multiple nodes, where each node is responsible for handling a specific stream object.
- The Query manager processes user-defined queries to detect and identify outliers in the streaming data. It utilizes the ISB information to determine the anomalous behaviour of the incoming samples and effectively captures the evolving stream patterns.

### 3.6 Kitsunes online algorithm

Kitsunes online algorithm (Kitnet) [25] is a light-weight neural network based outlier detection algorithm designed for identifying outliers in streaming data. Its architecture is optimized for low computational complexity, making it highly efficient for real-time anomaly detection. KitNet is composed of four primary components:

- **Packet capturer:** It captures the incoming streams and monitors the real-time traffic.
- **Packet parser:** It extracts the meta data from the incoming stream thereby helping in structuring the received in original data for further processing.
- **Feature extractor:** Here, auto encoder (AE) is employed to extract relevant features from data stream. This acts as dimensionality reduction, while preserving the important information.
- **Anomaly detector:** It analyzes the extracted features and classifies the data point either as normal or outliers.

### 3.7 K nearest neighbor conformal and density estimation method

K nearest neighbor conformal and density estimation method (KNN CAD) [26] is an ensemble-based outlier detection method that integrates two key anomaly detection approaches, i.e., (i) conformalized density-based anomaly detection and (ii) distance-based anomaly detection. Through these approaches, KNN-CAD utilizes both density and distance-based techniques, allowing it to effectively identify anomalies in streaming data. The anomaly score for each incoming stream object is computed based on previously processed data, ensuring that the method adapts dynamically to evolving patterns. Further, it incorporates a probabilistic interpretation of anomaly score, achieved through the conformal paradigm. Making it more robust in estimating deviations from normal behaviour.

## 4. Proposed Methodology

We present our proposed framework (refer to Algorithm 1) in a step-wise manner, and the corresponding schematic diagram is presented in Fig. 1. The framework is designed to handle both static (offline) model training and real-time incremental learning, making it suitable for dynamic, streaming environments. Further, we also depicted the flowchart in Fig. 2.

The proposed framework is broadly categorized into two stages:

(i) **Stage-I:** ***Offline model building*** The model is trained using thus available historical data in static settings, and

(ii) **Stage-II:** ***Incremental Learning in real-time*** The pre-trained model is deployed in streaming settings, where it is continuously updated as and when new data arrives.

**Stage-I: Offline model building in static settings**

This stage focuses on training a baseline model using the available historical data before transitioning to real-time learning.

1. **Step-1:** ***Initialization*** The first most step of this framework is to initialize the required number of parameters such as, model specific parameters, tracking variables such as time step counter to trace the time step (see lines 1-3, Algorithm 1).
2. **Step-2:** ***Collecting the training data***
   **a.** Since, this stage operates in a static environment, complete historical data is reserved as training data.
   **b.** This dataset is used for model training, allowing the model to learn initial relationships, data patterns, correlations etc.
   **c.** Feature and target variable information is stored in their respective variables.

3. **Step-3:** ***Model building in static settings***
   **a.** By utilizing the data collected in Step 2, we need to build the model and fine-tune it by using the grid search-based hyperparameter tuning technique (see line 6, Algorithm 1).
   **b.** This also addresses the cold-start problem, i.e., the performance of the model in the initial phase will be relatively lower until the model is trained with sufficient data. With this step, we could tackle this problem effectively (see lines 4-5, Algorithm 1).

After following the above three steps, the trained model is deployed in real-time settings, where it is continuously adapted to new information by following an incremental learning approach.

**Stage-II: Building model with incremental technique in streaming settings**

Unlike Stage-I, where the model is trained on the fixed historical dataset, Stage-II introduces a continuous learning mechanism that allows the model to dynamically adapt to incoming data streams. This makes the model stay up-to-date and robust while handling complex scenarios.

4. **Step-4:** ***Collecting First stream*** Since the environment now operates in streaming settings, new data arrives in stream after another. In this step, the first data stream is collected and stored in the respective storage.
5. **Step-5:** ***Incremental technique***
   **a.** In this step, the model that was built in Step 3 of Stage-I is now built in an incremental way by employing a partial training method.
   **b.** The partial training method refers to updating the model learned weighted in limited epochs, thereby not completely modifying previously learned patterns.
   **c.** In this way, instead of retraining the entire model from the beginning, only the new data is employed to fine-tune the existing model.

6. **Step-6:** ***Increment time step counter*** Now, the time step counter needs to be incremented with step size 1.
7. **Step-7:** ***Continous learning*** The above steps of collecting the data and incrementally building the model, thereafter incrementing the counter, is repeated until all the streams are exhausted.

**Algorithm 1: Proposed hybrid framework with incremental technique**

**Input:** Training data, $M_t$: Model built at $t^{th}$ time step.
**Output:** $M_{t+1}$, AUC

*// Stage-1: Offline model building with the available data*

1. k ← 1
2. t ← 0
3. $M_t$ ← Initialize the model with the respective parameter
4. X_train ← Store features information from Training data
5. y_train ← Store the class label from Training data
6. $M_t$ ← Train the model with X_train, y_train

*// Stage-2: Employed in real-time settings to train the model*

7. $S_t$ ← getStream()
8. **While** len($S_t$) > 0 **do**
9. data ← Collect the stream data
10. $M_{t+1}$ ← Partially fit the model with data
11. t ← t+1
12. $S_t$ ← getStream()
13. **End While**
14. AUC ← Compute results
15. Return $M_{t+1}$

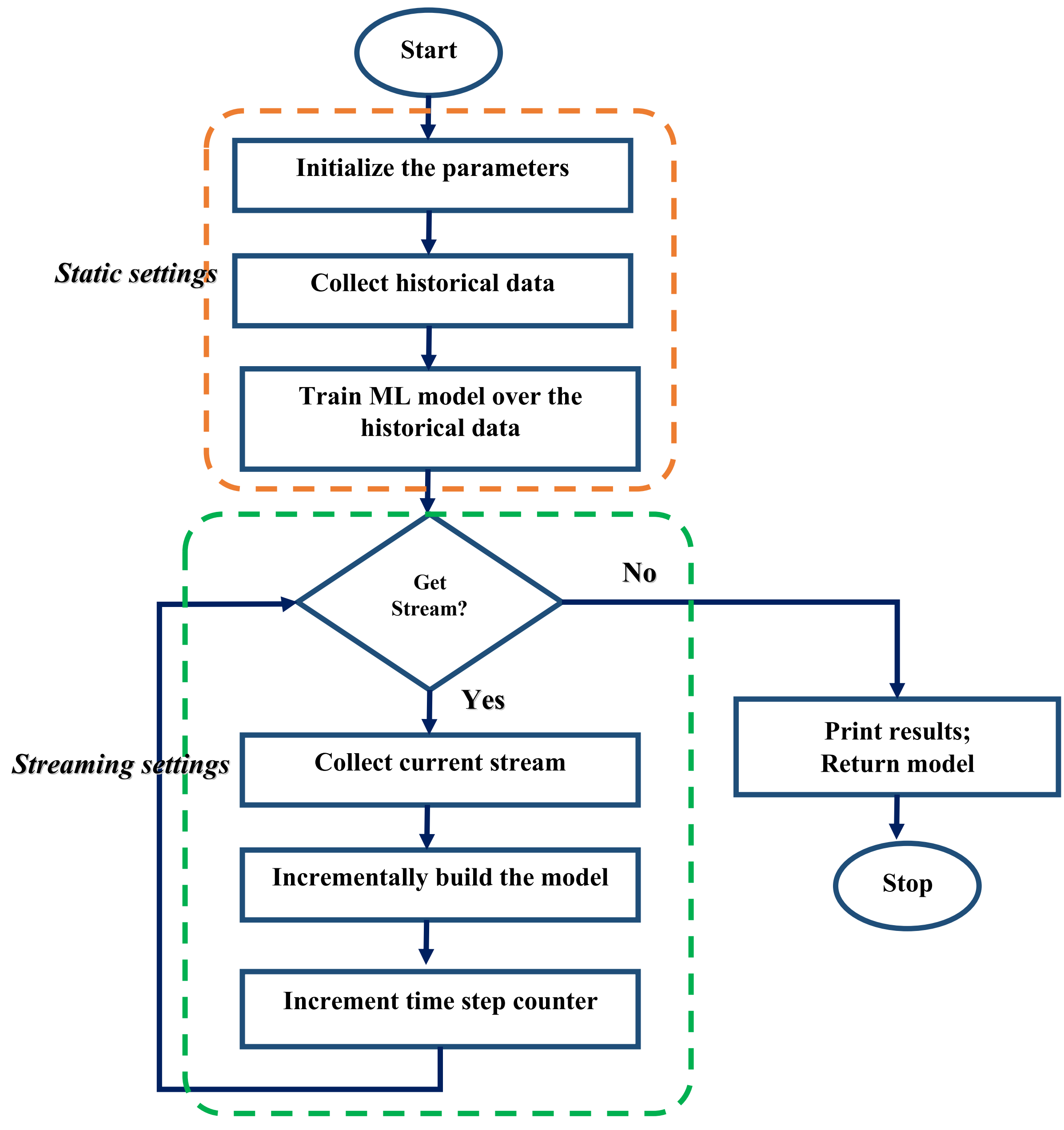

**Fig. 2. Flowchart of the proposed hybrid framework**

**Table 1: Description of the datasets**

| Dataset | #Samples | #Features | Class distribution |
|---|---|---|---|
| | | | Negative: Positive |
| **Credit card Churn prediction dataset [30]** | 10,000 | 13 | 80:20 |
| **Default of Credit Card Clients Dataset [30]** | 30,000 | 24 | 78:22 |
| **Auto insurance claims fraud dataset [30]** | 1,000 | 40 | 75:25 |
| **Ethereum fraud detection dataset [30]** | 9,841 | 51 | 86.3:13.7 |
| **Diabetes prediction dataset [29]** | 981 | 12 | 55 : 45 |
| **Brain stroke prediction dataset [29]** | 4,981 | 11 | 95: 5 |
| **Heart failure prediction dataset [29]** | 368 | 61 | 74.8: 25.2 |

# 5. Results & Discussion

## 5.1 Environmental Setup

The configuration of the experimental study conducted is as follows: i5 8$^{th}$ generation, octa-core, 2.4 GHz. The python version used for the experimental study is python 3.8 and we set up an anaconda environment to work on the jupyter notebook. All the experiments under a similar environment.

## 5.2 Dataset description

We considered seven benchmark datasets, and the corresponding metadata is presented in Table 1. These datasets span two domains – four from finance and three from health care domains. All these datasets are binary classification datasets, with 6 out of 7 datasets being highly imbalanced datasets. To simulate a streaming environment, we employed the pysade framework. We maintained the proportion as 30:70 for the static and streaming settings, i.e., 30% of the data was designated as the historical dataset for training the ML models under static settings, while the remaining 70% was reserved for the real-time environment.

## 5.3 Evaluation Metric

Area under receiver operating characteristic curve (AUC) is chosen to be the evaluation metric. While handling imbalanced datasets, AUC is proven to be one of the robust metric. It is calculated as an average of specificity and sensitivity (see Eq. (1)).

$$AUC = \frac{(Sensitivity+Specificity)}{2} \quad (1)$$

Where,

$$Sensitivity = \frac{TP}{TP+FN} \text{ and} \quad (2)$$

$$Specificity = \frac{TN}{TN+FP} \quad (3)$$

where TP is a true positive, FN is a false negative, TN is a true negative, and FP is a false positive.

## 5.4 Discussion over AUC

In this subsection, we compare various streaming OD models such as, OCSVM, IForest ASD, ES, ABOD, LOF, KitNet, and KNN CAD. The results are presented in Table 2-8 for all the benchmark datasets. To see the effectiveness of incremental training, we conducted the analysis without training in real-time settings. Hence, the experiment analysis is conducted in two different scenarios.

- **Scenario 1: Without incremental training:** This framework follows the static settings, but in the streaming settings, incremental training does not happen.
- **Scenario 2: With incremental training:** This framework follows the proposed methodology both in static and streaming settings.

From the results (see Tables 2-8), it is clearly evident that models in scenario 2, i.e., after incorporating incremental learning approach consistently outperformed their scenario 1 counterparts in terms of AUC. This highlights the importance of retraining with new data streams to maintain the model generalization. Most importantly, this observation is noticed despite handling the cold start problem in scenario 1 by training the models with the historical data. The superior performance of scenario 2 is attributed to several factors, such as:

- Incorporating incremental learning allows models to continuously update their knowledge, making them more responsive to evolving streaming patterns.
- Scenario 2 is better equipped to handle abrupt data shifts effectively, thereby retraining the model to improve its generalization ability.

These factors collectively demonstrate why models trained incrementally outperform static, offline-trained models in dynamic environments. Additionally, in 5 out of 7 datasets, except the default of credit card clients and diabetes prediction datasets, the top-performing model is achieved from the models trained in scenario 2.

While comparing models, the following are the important observations made:

- IForest ASD algorithm turned out to be the best-performing algorithm across all datasets in 4 out of 7 datasets. This is due to the ensemble nature of the model, which makes it a better generalization model and makes it robust to handle abrupt data shift effectively than other models.
- ABOD turned out to be the second-best algorithm, where it turned out to be the best in the brain stroke dataset by obtaining an AUC of 0.751. Further, it consistently ranked in top 3 across different datasets. ABOD models are generally very good for low-dimensional datasets. However, as the number of features increases, the performance of ABOD is decreased.
- Exact storm also performed equally well to ABOD in the majority of the cases and turned out to be the best-performing algorithm in the diabetes dataset by obtaining an AUC of 0.621. However, it failed to handle the abrupt data shift effectively, thereby decreasing the performance of the model.
- Except in the Ethereum dataset, KitNet failed miserably when compared to the top-performing algorithm. This is due to the fact that KitNet requires a larger number of data samples to improve its generalization ability.
- LOF performance across the dataset is okay but could only obtain second position in four datasets, i.e., ethereum dataset, credit card fraud dataset, diabetes and brain stroke datasets. This shows that it effectively identified the local patterns of outliers.

- OCSVM and KNN CAD are the worst-performing algorithms across all datasets. This is mainly likely due to the former OCSVM being computationally highly expensive and could not scale well to high dimensional or voluminous data. Further, KNN CAD is too simple and non-parametric method, failing the model to extract meaningful patterns as and when either the number of samples or dimensions increases.

In summary the following are the key observations:

- Models trained under framework with incremental learning is the best in 5 out of 7 datasets.
- Among the models, IForest turned out to be the best performing algorithm in 4 out of 7 datasets.

**Table 2: AUC obtained in credit card fraud detection dataset**

| Name of the model | Proposed hybrid framework | |
|---|---|---|
| | Without incremental learning | With incremental learning |
| ABOD | 0.518 | 0.571 |
| LOF | 0.469 | 0.570 |
| OCSVM | 0.5 | 0.5 |
| IForest ASD | 0.496 | **0.611** |
| Exact Storm | 0.5 | 0.462 |
| KitNet | 0.5 | 0.546 |
| KNN CAD | 0.5 | 0.489 |

**Table 3: AUC obtained in Ethereum fraud detection dataset**

| Name of the model | Proposed hybrid framework | |
|---|---|---|
| | Without incremental learning | With incremental learning |
| ABOD | 0.503 | 0.579 |
| LOF | 0.432 | 0.625 |
| OCSVM | 0.544 | 0.475 |
| IForest ASD | 0.336 | 0.527 |
| Exact Storm | 0.593 | 0.508 |
| KitNet | 0.5 | **0.668** |
| KNN CAD | 0.5 | 0.501 |

**Table 4: AUC obtained in Auto insurance claims dataset**

| Name of the model | Proposed hybrid framework | |
|---|---|---|
| | Without incremental learning | With incremental learning |
| ABOD | 0.487 | 0.490 |
| LOF | 0.436 | 0.486 |
| OCSVM | 0.5 | 0.5 |
| IForest ASD | 0.471 | **0.589** |
| Exact Storm | 0.5 | 0.474 |
| KitNet | 0.5 | 0.5 |
| KNN CAD | 0.5 | 0.5 |

**Table 5: AUC obtained in Default of credit card clients dataset**

| Name of the model | Proposed hybrid framework | |
|---|---|---|
| | Without incremental learning | With incremental learning |
| ABOD | 0.380 | 0.546 |
| LOF | 0.410 | 0.556 |
| OCSVM | 0.498 | 0.5 |
| IForest ASD | **0.595** | 0.571 |
| Exact Storm | 0.5 | 0.536 |
| KitNet | 0.5 | 0.428 |
| KNN CAD | 0.5 | 0.503 |

**Table 6: AUC obtained in Diabetes prediction dataset**

| Name of the model | Proposed hybrid framework | |
|---|---|---|
| | Without incremental learning | With incremental learning |
| ABOD | 0.502 | 0.572 |
| LOF | 0.380 | 0.607 |
| OCSVM | 0.468 | 0.5 |
| IForest ASD | 0.412 | 0.475 |
| Exact Storm | **0.621** | 0.523 |
| KitNet | 0.5 | 0.5 |
| KNN CAD | 0.5 | 0.4 |

**Table 7: AUC obtained in Brain stroke dataset**

| Name of the model | Proposed hybrid framework | |
|---|---|---|
| | Without incremental learning | With incremental learning |
| ABOD | 0.580 | **0.751** |
| LOF | 0.490 | 0.725 |
| OCSVM | 0.537 | 0.5 |
| IForest ASD | 0.631 | 0.557 |
| Exact Storm | 0.5 | 0.409 |
| KitNet | 0.5 | 0.478 |
| KNN CAD | 0.5 | 0.524 |

**Table 8: AUC obtained in Heart failure prediction dataset**

| Name of the model | Proposed hybrid framework | |
|---|---|---|
| | Without incremental learning | With incremental learning |
| ABOD | 0.39 | 0.41 |
| LOF | 0.49 | 0.413 |
| OCSVM | 0.405 | 0.5 |
| IForest ASD | 0.439 | **0.528** |
| Exact Storm | 0.5 | 0.447 |
| KitNet | 0.5 | 0.5 |
| KNN CAD | 0.5 | 0.5 |

# 6. Conclusion and Future Work

In this work, the proposed framework with incremental learning was evaluated against offline-trained models to assess the effectiveness of incremental learning within a streaming framework. The results clearly demonstrate that incremental models outperform offline models, particularly in scenarios involving highly imbalanced datasets, where the majority class significantly outweighs the minority class. This highlights the advantage of continuous learning in environments where data distribution evolves. Among all the models tested, the ensemble-based IForest ASD model consistently delivered superior performance. Its ability to dynamically adapt to new data and handle abrupt shifts in data distribution makes it particularly well-suited for real-world applications involving anomaly detection. These findings reinforce the necessity of employing incremental learning approaches in streaming data environments to ensure models remain accurate, adaptive, and robust against abrupt data shifts.

In Future work, a robust model has to be designed by improving the performance of the models. The robustness can be increased by ensembling various models, thereby enhancing the voting-based mechanism, which alleviates the advantages of various models.

# References

1. G.B. Gebremeskel, C. Yi, Z. He, D. Haile. Combined data mining techniques based patient data outlier detection for healthcare safety. *International Journal of Intelligent Computing and Cybernetics*, *9*(1), (2016), 42-68.
2. M. Ahmed, A.N. Mahmood, M. R. Islam. A survey of anomaly detection techniques in financial domain. *Future Generation Computer Systems* 55 (2016), 278-288.
3. P. Najafi, F. Cheng, C. Meinel. HEOD: Human-assisted Ensemble Outlier Detection for cybersecurity. *Computers & Security*, *146*, 104040. (2024). https://doi.org/10.1016/j.cose.2024.104040
4. I. Souiden, M.N. Omri, Z. Brahmi. . A survey of outlier detection in high dimensional data streams. *Computer Science Review*, *44*, 100463. (2022). https://doi.org/10.1016/j.cosrev.2022.100463
5. V. Barnett, T. Lewis. Outliers in Statistical Data. John Wiley & Sons, 3 edition (1993).
6. N. Malini, M. Pushpa. Analysis on credit card fraud identification techniques based on KNN and outlier detection. In2017 third international conference on advances in electrical, electronics, information, communication and bio-informatics (AEEICB) (2017), pp. 255-258.
7. Y. Vivek, V. Ravi, P.R. Krishna. Online feature subset selection for mining feature streams in big data via incremental learning and evolutionary computation. *Swarm and Evolutionary Computation*, *94*, 101896, (2025). https://doi.org/10.1016/j.swevo.2025.101896
8. H. Yuan, A.A. Hernandez. User cold start problem in recommendation systems: A systematic review. IEEE access. Dec 1;11:136958-77 (2023).
9. A. Degirmenci, O. Karal. IMCOD: Incremental multi-class outlier detection model in data streams. *Knowledge-Based Systems*, *258*, 109950. (2022) https://doi.org/10.1016/j.knosys.2022.109950
10. Y. Yang, C. Fan, L. Chen, H. Xiong. IPMOD: An efficient outlier detection model for high-dimensional medical data streams. *Expert Systems With Applications*, *191*, 116212, (2022). https://doi.org/10.1016/j.eswa.2021.116212
11. S. Cai, L. Li, S. Li, R. Sun, G. Yuan. An efficient approach for outlier detection from uncertain data streams based on maximal frequent patterns. *Expert Systems With Applications*, *160*, 113646, (2020). https://doi.org/10.1016/j.eswa.2020.113646
12. J. Gao, W. Ji, L. Zhang, A. Li, Y. Wang, Z. Zhang. Cube-based incremental outlier detection for streaming computing. *Information Sciences*, *517*, (2020), 361-376.

https://doi.org/10.1016/j.ins.2019.12.060

13. W. Xiaolan, M. M. Ahmed, N.N. Husen, Z. Qian, S. B. Belhaouari. Evolving anomaly detection for network streaming data. *Information Sciences*, *608*, (2022), 757-777. https://doi.org/10.1016/j.ins.2022.06.064
14. A. Raut, A. Shivhare, V.K. Chaurasiya, M. Kumar. AEDS-IoT: Adaptive clustering-based Event Detection Scheme for IoT data streams. *Internet of Things*, *22*, 100704. (2023). https://doi.org/10.1016/j.iot.2023.100704
15. K. Ducharlet, L. Travé-Massuyes, J.B. Lasserre, M.V. Lann, Y. Miloudi. Leveraging the Christoffel function for outlier detection in data streams. *Int J Data Sci Anal* (2024). https://doi.org/10.1007/s41060-024-00581-2
16. Z. Hu, X. Yu, L. Liu, H. Yu. ASOD: an adaptive stream outlier detection method using online strategy. *J Cloud Comp* **13**, 120 (2024). https://doi.org/10.1186/s13677-024-00682-0
17. S. Pei, J. Wen, K. Xie, G. Xie and K. Li, "On-Line Network Traffic Anomaly Detection Based on Tensor Sketch," in IEEE Transactions on Parallel and Distributed Systems, vol. 34, no. 12, pp. 3028-3045, (2023), doi: 10.1109/TPDS.2023.3316717.
18. O. Alghushairy, R. Alsini, Z. Alhassan, A.A. Alshadadi, A. Banjar, A. Yafoz. An Efficient Support Vector Machine Algorithm Based Network Outlier Detection System," in IEEE Access, vol. 12, pp. 24428-24441, (2024), doi: 10.1109/ACCESS.2024.3364400.
19. Y. Devavarapu, R. R. Bedadhala, S. S. Shaik, C. R. K. Pendela and K. Ashesh. "Credit Card Fraud Detection Using Outlier Analysis and Detection," 2024 4th International Conference on Intelligent Technologies (CONIT), Bangalore, India, (2024), pp. 1-7, doi: 10.1109/CONIT61985.2024.10626480.
20. M. J. Bah, H. Wang, M. Hammad, F. Zeshan and H. Aljuaid, "An Effective Minimal Probing Approach With Micro-Cluster for Distance-Based Outlier Detection in Data Streams," in IEEE Access, vol. 7, pp. 154922-154934, (2019), doi: 10.1109/ACCESS.2019.2946966.
21. S.F. Yilmaz, S.S. Kozat. PySAD: A Streaming Anomaly Detection Framework in Python. *ArXiv*. (2020). https://arxiv.org/abs/2009.02572
22. F. Angiulli, F. Fassetti. Detecting distance-based outliers in streams of data. In *Proceedings of the sixteenth ACM conference on Conference on information and knowledge management*, 811–820. (2007).
23. H-P. Kriegel, P. Kroger, E. Schubert, A. Zimek. Loop: local outlier probabilities. In *Proceedings of the 18th ACM conference on Information and knowledge management*, 1649–1652. (2009).
24. Z. Ding, M. Fei. An anomaly detection approach based on isolation forest algorithm for streaming data using sliding window. *IFAC Proceedings Volumes*, 46(20):12–17, (2013).
25. Y. Mirsky, T. Doitshman, Y. Elovici, A. Shabtai. Kitsune: an ensemble of autoencoders for online network intrusion detection. *arXiv preprint arXiv:1802.09089*, 2018 (2018).
26. E. Burnaev, V. Ishimtsev. Conformalized density-and distance-based anomaly detection in time-series data. *arXiv preprint arXiv:1608.04585*, (2016).
27. B. Schölkopf, J.C. Platt, J. S-Taylor, A.J. Smola, R.C. Williamson. Estimating the support of a high-dimensional distribution. Neural computation, 13(7):1443–1471, (2001).
28. H-P. Kriegel, A. Zimek et al. Angle-based outlier detection in high-dimensional data. In Proceedings of the 14th ACM SIGKDD international conference on Knowledge discovery and data mining, 444–452. ACM, (2008).
29. UCI repository: https://archive.ics.uci.edu/ml/index.php
30. Kaggle repository: https://www.kaggle.com/